\documentclass{article}

\usepackage{times}
\usepackage{graphicx} 
\usepackage{subfigure} 
\usepackage{tabularx}
\usepackage{tabulary}
\usepackage{natbib}
\usepackage{multirow}
\usepackage[T1]{fontenc}

\usepackage{algorithm}
\usepackage{algorithmic}
\usepackage{amsmath}
\usepackage{amssymb}
\usepackage{color}
\usepackage{amsthm}
\usepackage[bookmarks=false]{hyperref}
\usepackage{comment}

\usepackage[accepted,nohyperref]{icml2016}

\begin{document} 

\twocolumn[
\icmltitle{Recurrent Orthogonal Networks and Long-Memory Tasks}

\icmlauthor{Mikael Henaff}{mbh305@nyu.edu}
\icmladdress{New York University, Facebook AI Research}
\icmlauthor{Arthur Szlam}{aszlam@facebook.com}
\icmladdress{Facebook AI Research}
\icmlauthor{Yann LeCun}{yann@cs.nyu.edu}
\icmladdress{New York University, Facebook AI Research}

\icmlkeywords{boring formatting information, machine learning, ICML}

\vskip 0.3in
]

\begin{abstract} 
Although RNNs have been shown to be powerful tools for processing sequential data, finding architectures or optimization strategies that allow them to model very long term dependencies is still an active area of research.   
In this work, we carefully analyze two synthetic datasets originally outlined in \cite{Hochreiter1997} which are used to evaluate the ability of RNNs to store information over many time steps.  
We explicitly construct RNN solutions to these problems, and using these constructions,  illuminate both the problems themselves and the way in which RNNs store different types of information in their hidden states. 
These constructions furthermore explain the success of recent methods that specify unitary initializations or constraints on the transition matrices. 
\end{abstract} 

\section{Introduction}



Recurrent Neural Networks (RNNs) are powerful models which are naturally suited to processing sequential data. 
They can maintain a hidden state which encodes information about previous elements in the sequence. 
For a classical version of RNN \cite{Elman90findingstructure}, at every timestep, the hidden state is updated as a function of both the input and the current hidden state. 
In theory, this recursive procedure allows these models to store complex signals for arbitrarily long timescales. 

However, in practice RNNs are considered difficult to train due to the so-called vanishing and exploding gradient problems \cite{Bengio94}.
These problems arise when the spectral norm of the transition matrix is significantly different than 1, or due to the non-linear transition functions. 
If the spectral norm of the transition matrix is greater than 1, the gradients will grow exponentially in magnitude during backpropagation, which is known as the exploding gradient problem. 
If the spectral norm is less than 1, the gradients will vanish exponentially quickly, which is known as the vanishing gradient problem. 
Recently, a simple strategy of clipping gradients has been introduced, and has proved effective in addressing the exploding gradient problem \cite{mikolov_thesis}. 
The problem of vanishing gradients has shown itself to be more difficult, and various strategies have been proposed over the years to address it. 
One very successful approach, known as Long Short-Term Memory (LSTM) units \cite{Hochreiter1997}, has been to modify the architecture of the hidden units by introducing gates which explicitly control the flow of information as a function of both the state and the input. 
Specifically, the signal stored in a hidden unit must be explicitly erased by a forget gate and is otherwise stored indefinitely. 
This allows information to be carried over long periods of time.   LSTMs  have become very successful in applications to language modeling, machine translation, and speech recognition \cite{Zaremba2014, Sutskever14, Graves2013b}. 
Other methods have been proposed to deal with learning long-term dependencies, such as adding a separate contextual memory \cite{Mikolov2015}, stabilizing activations \cite{Krueger2015} or using more sophisticated optimization schemes \cite{Martens2011}. 
Two recent methods propose to directly address the vanishing gradient problem by either initializing or parameterizing the transition matrix with orthogonal or unitary matrices \cite{Arjovsky2015, Le2015}.

These works have used a set of synthetic problems (originally outlined in \cite{Hochreiter1997} or variants thereof) for testing the ability of methods to learn long-term dependencies. 
These synthetic problems are designed to be pathologically difficult, and require models to store information over very long timescales (hundreds of timesteps). 
Different approaches have solved these problems to varying degrees of success. 
In \cite{Martens2011}, the authors report that their Hessian-Free optimization based method solves the addition task for $T=200$ timesteps. 
The authors of ~\cite{Krueger2015} reported that their method beat the chance baseline for the adding task in 8/9 cases for $T=400$.
In \cite{Le2015}, the IRNN is reported to solve the addition task for $T=300$. 
The method proposed in \cite{Arjovsky2015} is able to solve the copy task for up to $T=500$ timesteps, and is able to completely solve the addition task for up to $T=400$ timesteps, and partially solves it for $T=750$. 

In this work we analyze these ``long-memory'' tasks, and construct explicit RNN solutions for them.  
The solutions illuminate both the tasks, and provide a theoretical justification for the success of recent approaches using orthogonal initializations of, or unitary constraints on, the transition matrix of the RNN. 
In particular, we show that a classical Elman RNN with no transition non-linearity and random orthogonal initialization is with high probability close to our explicit RNN solution to the sequence memorization task, and the same network architecture with identity initialization is close to the explicit solution to the addition task.   We verify experimentally that initializing correctly (i.e. random orthogonal or identity) is critical for success on these tasks.
Finally, we show how $l_2$ pooling can be used to allow a model to ``choose'' between a random-orthogonal or identity-like memory. 

There are several other works which have studied the properties of orthogonal matrices in relation to neural networks.
The work of \cite{saxe2013} gives exact solutions to the learning dynamics of deep linear networks, and based on this analysis, suggests an orthogonal initialization scheme to accelerate learning.
The authors of \cite{White2004} and \cite{Ganguli2008} study the ability of linear RNNs (with orthogonal and generic transition matrices, respectively) to store scalar sequences in their hidden state, and show that the memory capacity scales with the number of hidden units.  
Our work complements theirs by providing a related analysis for discrete input sequences.

\section{Architectures}

We review some recurrent neural network (RNN) architectures for processing sequential data, and discuss the modifications we use for the long memory problems.  
 We fix the following notation: input sequences are denoted
$x_0,\, x_1,\, ... , x_t, ...$, and output sequences are denoted by $y_0,\, y_1,\, ... , y_t, ...$ .

\subsection{sRNN}
An sRNN \cite{Elman90findingstructure}  consists of a $d \times d$ transition matrix $V$, an $M\times d$ decoder matrix $W$ (where $M$ is the output dimension),  a $d \times N$ encoder matrix $U$ (where $N$ is the input dimension), and a bias $b$.   
If either the output or input is categorical, $M$ (respectively $N$) is the number of classes, and we use a one-hot representation.   As the sRNN ingests a sequence, it keeps running updates to a hidden state $h$, and using the hidden state and the decoder matrix, produces outputs $y$:
\begin{equation}
\label{eq:elman}
\begin{split}
h_t &= \sigma(Ux_t + Vh_{t-1} + b) \\
y_t &= Wh_t
\end{split}
\end{equation}
where $x_t, y_t, h_t$ are the input, output and hidden state respectively at time $t$.   
While there have been great improvements in the training of sRNNs since their introduction, and while they have been shown to be powerful models in tasks such as language modeling \cite{mikolov_thesis}, it can still be difficult to train generic sRNNs to use information about inputs from hundreds of timesteps previous for computing the current output \cite{Bengio94, Pascanu13}.

In the following, we will use a simplification of the sRNNs that makes them in some sense less powerful models, but makes it easier to train them to solve simple long-memory tasks.  Namely, by placing the non-linearity between the input and hidden state, rather than between the hidden state and output, we obtain RNNs with linear transitions (or, in the case of categorical inputs, not using a non-linearity at all). 
We call these LT-RNNs. 
The update equations are then: 
\begin{equation}
\label{eq:LTRNN}
\begin{split}
h_t &= \sigma(Ux_t + b) + Vh_{t-1} \\
y_t &= Wh_t
\end{split}
\end{equation}
Finally, note that by appropriately scaling the weights and biases, an sRNN can be made to approximate a LT-RNN, but not the other way around (and of course the optimization may never find this scaling).
\subsection{LSTM}
The LSTM of \cite{Hochreiter1997} is an architecture designed to improve upon the sRNN with the introduction of simple memory cells with a gating architecture. 
In this work we use the architecture originally proposed in \cite{Hochreiter1997}. 
For each memory cell, the network computes the output of four gates: an update gate, input gate, forget gate and output gate. 
The outputs of these gates are:

\begin{equation}
\begin{split}
i &= \sigma(U_ix_t + V_ih_{t-1}) \\
f &= \sigma(U_fx_t + V_fh_{t-1}) \\
o &= \sigma(U_ox_t + V_oh_{t-1}) \\
g &= \mbox{tanh}(U_gx_t + V_gh_{t-1}) \\
\end{split}
\end{equation}

The cell state is then updated as a function of the input and the previous state:

\begin{equation}
c_t = f \odot c_{t-1} + i \odot g
\end{equation}

Finally, the hidden state is computed as a function of the cell state and the output gate:

\begin{equation}
h_t = o \odot \mbox{tanh}(c_t)
\end{equation}

A relatively common variation of the original LSTM involves adding so-called ``peephole connections'' \cite{Gers2003} which allows information to flow from the cell state to the various gates. 
This variant was originally designed to measure or generate precise time intervals, and has proven successful for speech recognition and sequence generation \cite{Graves2013a, Graves2013b}. 

\subsection{LT-RNN with $l_2$ pooling}
\label{sec:pooling}
Below we will consider LT-RNNs initialized with either random orthogonal transition matrices, or identity transitions, and we will see that there is a large difference in behavior between these initializations.  However, we can set up an architecture where a random orthogonal initialization behaves  much closer to an identity initialization by using an $l_2$ pooling layer at the output.   If we feed both the pooled and unpooled hidden layer to the decoder, the model can choose whether it wants an identity-like or random-orthogonal like representation.  We fix a pool size $k$, and then the update equations for this  model  are:
\begin{equation}
\label{eq:pooled_RNN}
\begin{split}
h_t &= \sigma(Ux_t + b) + Vh_{t-1} \\
y_t &= W_{I}h_t + W_P P_k(h_t)
\end{split}
\end{equation}
where if $h$ is the $kd$ dimensional vector $h = [h_1 , ... , h_{kd}]^T$, then $P(h)$ is the $d$ dimensional vector defined by 
\[P(h)_i = \sqrt{\sum_{j = k(i-1) +1}^{ki} h_j^2}\]

\section{Tasks}
In this section we describe tasks from \cite{Hochreiter1997,Arjovsky2015,Le2015} which involve dependencies over very long timescales which are designed to be pathologically hard for the sRNN. 
\subsection{Copying Problem}

This task tests the network's ability to recall information seen many time steps previously. 
We follow the same setup as \cite{Arjovsky2015}, which we briefly outline here. 
Let $A=\{a_i\}_{i=1}^K$ be a set of $K$ symbols, and pick numbers $S$ and $T$. 
The input consists of a $T + 2S$ length vector of categories, starting with $S$ entries sampled uniformly from $\{a_i\}_{i=1}^{K}$ which are the sequence to be remembered. 
The next $T - 1$ inputs are set to $a_{K+1}$, which is a blank category. 
The following (single) input is $a_{K+2}$, which represents a delimiter indicating that the network should output the initial $S$ entries of the input. 
The last $S$ inputs are set to $a_{K+1}$. 
The required output sequence consists of $T+S$ entries of $a_{K+1}$, followed by the first $S$ entries of the input sequence in exactly the same order. 
The task is to minimize the average cross-entropy of the predictions at each time step, which amounts to remembering a categorical sequence of length $S$ for $T$ time steps. 

\subsubsection{a solution mechanism}
\label{sec:copy_solution}
We can write out an LT-RNN solution for this problem. We will write out descriptions for $U,V,W$ from equation \eqref{eq:LTRNN}.
Note that since the inputs are categorical, we assume that no non-linearity is used.
Fix a number $d$.     For each $j$ in $\{1,...,d\}$, pick a random integer $l_j$ drawn uniformly from $\{ 1, ... ,T+S\}$, and let 
\[Q_j = \begin{pmatrix}  \cos (2l_j\pi/(T+S)) & \sin (2l_j\pi/(T+S)) \\
						 -\sin (2l_j\pi/(T+S)) & \cos (2l_j\pi/(T+S))
		\end{pmatrix}.
		\]
Now define  $Q$,   and then $V$ from \eqref{eq:LTRNN} by  
        \[Q = \begin{pmatrix} Q_1     & 0    & \cdots & 0      \\
 		                      0       & Q_2  & \cdots & 0      \\
 		                      \vdots	 &      & \ddots  & 0        \\
 		                      0       &      &   0    &  Q_d   
		 	  \end{pmatrix}, \,\,\,\,
					 	 V = \begin{pmatrix} Q     & 0      \\
 		                                         0    &  1   
		 	  \end{pmatrix}.
	\] 
So $V$ is a  $(2d+1) \times (2d+1)$ block diagonal matrix.  Note that iterating $Q$ ``spins'' each of the $Q_i$ at different rates, but they all synchronize at multiples of $S+T$.  Thus  $Q$ acts as a ``clock'' with period $S+T$.
 Now set
$\tilde{U}$ to be a $2d \times K$ matrix with columns sampled uniformly from the unit sphere, 
and form $U$ by appending two zero columns to $\tilde{U}$ and then one extra row, with $-1/S$ for each entry between 1 and $K$, $-1$ for the $K+1$ entry, and $T+S+1$ for the $K+2$ entry.  Schematically, 


\[U = \begin{bmatrix} \tilde{U} & 0 & 0\\
	            -\frac{1}{S} & -1 & S+T+1\\
	            \end{bmatrix}.\] 


Finally, set $W = U^T$, except scale the $K+1$ column by $S+1$, zero out the $K+2$ column, and also zero out the entries below $\tilde{U}$.

This gives
 
 \[W = \begin{bmatrix} \tilde{U}^T & 0\\
		       0  & -(S+1) \\
	            0 & 0\\
	            \end{bmatrix}.\]

Now we will show how the RNN operates, starting with a high-level overview.
The last dimension of the hidden state divides the state space into two regions, one where the model outputs the blank symbol and the other 
where it outputs one of the first $K$ symbols in the dictionary.
The model begins in the first region and remains there until it encounters the delimiter symbol, which sends it into the second.
Now, the symbols in the input sequence are all encoded in the hidden state and a rotation is applied with each timestep.
A key result is that rotation by powers of $Q$ ``hides'' a symbol encoded in the hidden state, i.e. decorrelates its current representation from its original one. 
Due to the periodicity of $Q$, after $T + S$ timesteps, the different symbols in the input sequence will surface from the hidden state one at a time, in the order in which they were seen, 
while the other symbols in the sequence, whose representations have rotations applied to them, remain hidden. 
This causes the output units of each of the symbols to fire in the correct order.

We now give a more precise description, for which we need a little more notation.   
Denote by $\tilde{h}$ the first $2d$ coordinates of $h$, and by $h_{2d+1}$ the last coordinate, and denote by $u_j$ the $j$th column of $\tilde{U}$.  Then the RNN works as follows, initialized with hidden state $0$:
\begin{itemize}
	\item After the first $S$ inputs, we have 
		\[\tilde{h} = \sum_{j = 1}^S Q^{S-j} u_{i_j}.\]
	\item For the next $T$ inputs, only $h_{2d+1}$ changes, incrementing by $-1$ at each step.  Note that $W_{K+1}$ has so far been the best match to $h$ because it has large negative last component.
	\item At time $T+S$ when the $a_{K+2}$ token is seen, $h_{2d+1}$ is set positive, which ensures the blank symbol will not be output.
	\item At time $T+S+1$, \[\tilde{h} = u_{i_1} + \sum_{j = 2}^S Q^{1-j} u_{i_j}.\]  We argue below that 
		if $d$ is large enough w.r.t. $S$ and $K$, with high probability, $u_{i_1}^T\sum_{j = 2}^S Q^{1-j} u_{i_j}$ is small, and so multiplication with $W$ has max value at $i_1$. 
	\item The sum continues to cycle, giving $i_j$ as output for each following $j$ up to $S$.
\end{itemize}

\begin{figure}[t]
\centering 
\includegraphics[width=0.5\textwidth]{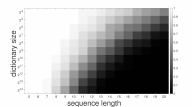}
\caption{Success percentage of the mechanism from \ref{sec:copy_solution} to the copy problem for $T=500$, computed over 500 trials with $d=128$.} 
\label{fig:copy-mechanism}
\end{figure}

We now briefly argue that $u_{i_1}^T\sum_{j = 2}^S Q^{1-j} u_{i_j}$ is small when $d$ is large enough w.r.t. $S$ and $K$.  We will repeatedly use that the variance of a sum of independent, mean-zero random variables grows
as the sum of the variances.
Denote by $u_{ji}$ the pair of coordinates of the $j$th column $u_j$ of $\tilde{U}$ corresponding to the $i^{th}$ block; since $u$ are uniform on the sphere, we expect 
\[||u_{ji}||^2 \sim 1/d.\]
  Since for each fixed $p\in \{1,T+S\}$, over the choices of $l_i$ in the definition of $Q_i$, the $Q_i^p$ are independent, $u_{ji}^TQ_i^pu_{ji}$ has mean zero, and since
\[
u_j^TQ^pu_{j} =\sum_{i=1}^d u_{ji}^TQ_i^pu_{ji}
\]
 we expect \[|u_{j}^TQ^pu_{j}|^2 \sim 1/d.\]  Moreover, since the $Q^pu_{j}$ are uniform on the sphere, \[|u_{j}^TQ^pu_{j'}|^2 \sim 1/d\] for $ j'\neq j$.  Similarly, we expect  
 \[\left|u_{i_1}^T\sum_{j = 2}^S Q^{1-j} u_{i_j}\right|^2 \sim (S-1)/d.\]  
 Thus we can fix a small number $\epsilon$, say $\epsilon = .1$, and choose $d$ large enough so that with high probability $|w_{i_1}^T\sum_{j = 2}^S Q^{j-S} u_{i_j}|^2< \epsilon$, even though $w_i^Tu_i = 1$.   Finally, there is a weak dependence on $K$ here; for fixed $\epsilon$ and $K$ it is exponentially unlikely (in $d$) that the nearest neighbor $i'\neq i_1 $ is close enough to $u_i$ to interfere.

This solution mechanism suggests that a random orthogonal matrix (chosen, for example, via QR decomposition of a Gaussian matrix) is a good starting point for solving this task.  The construction above is invariant to rotations; and we can always find a basis so that a given orthogonal matrix has the block form above in that basis.  Thus all that is necessary is for the descent to nudge the eigenvalues of the orthogonal matrix to be $S+T$ roots of unity, and then it already has the basic form of the construction above.  This also gives a good explanation for the performance of the models used for the copy problem in  \cite{Arjovsky2015}

Finally, note that although we used the setup of \cite{Arjovsky2015}, the construction can be modified to solve  problems 2a and 2b in \cite{Hochreiter1997}

\subsubsection{solution mechanism experiments}
Since the construction of the copy mechanism is randomized, we provide an experiment to show how the solution degrades as a function of $K$ (the dictionary size) and $S$ (the length of the sequence to be remembered).  There is not a strong dependence on $T$ (the length of time to remember the sequence).  Figure \ref{fig:copy-mechanism} shows the number of successes over $500$ runs with $d=128$.

\subsubsection{Variable Length Copy Problem}

 Note that the solution mechanism for the copy problem above depends on having a {\it fixed} location for regurgitating the input. In the experiments below, we also discuss a variant of the copy task, where the symbol to indicate that the memorized sequence must be output is randomly located in ${S + 1, S + T}$; this can be considered a variant of task 2c in \cite{Hochreiter1997}.   We do not know a bounded in $T$ explicit LT-RNN or sRNN solution for this variable length problem (although the above solution using a multiplicative RNN instead of an sRNN, and keeping $d^2$ 
extra hidden variables to track the power of $V$ solves it). 

\subsection{Adding Problem}

The adding problem requires the network to remember two marked numbers in a long sequence and add them. 
Specifically, the input consists of a two dimensional sequence $\{x_1, ... ,x_T\}$. 
The first coordinate $x_j[1$] is uniformly sampled between 0 and 1, and the second coordinate is 0 at each $j$ save two; in these two entries, $x_j[2] = 1$.  The required output is $x_{j_1}[1] + x_{j_2}[1]$, where $x_{j_i}[2] = 1$.

\begin{figure*}[t]
    \centering
    \subfigure{\includegraphics[width=0.95\columnwidth]{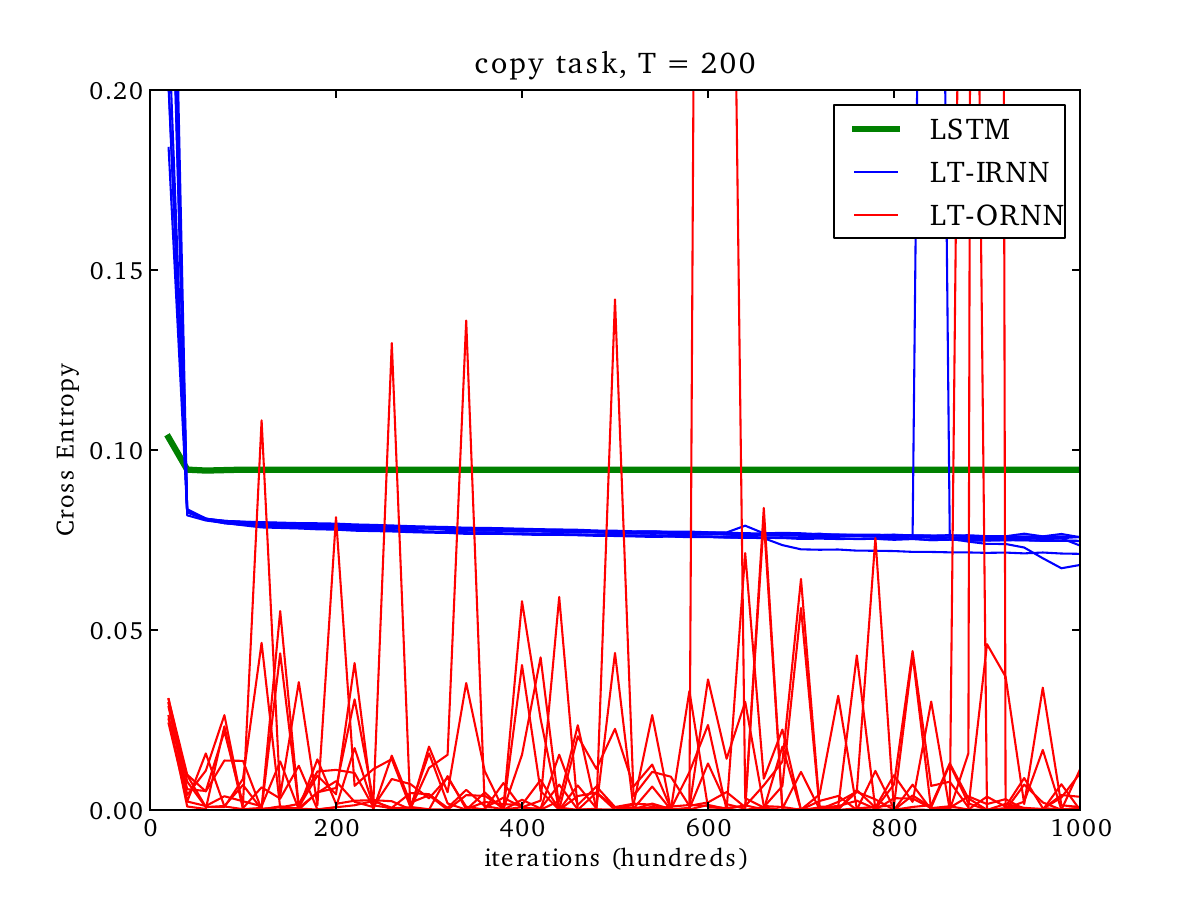}}\quad
    \subfigure{\includegraphics[width=0.95\columnwidth]{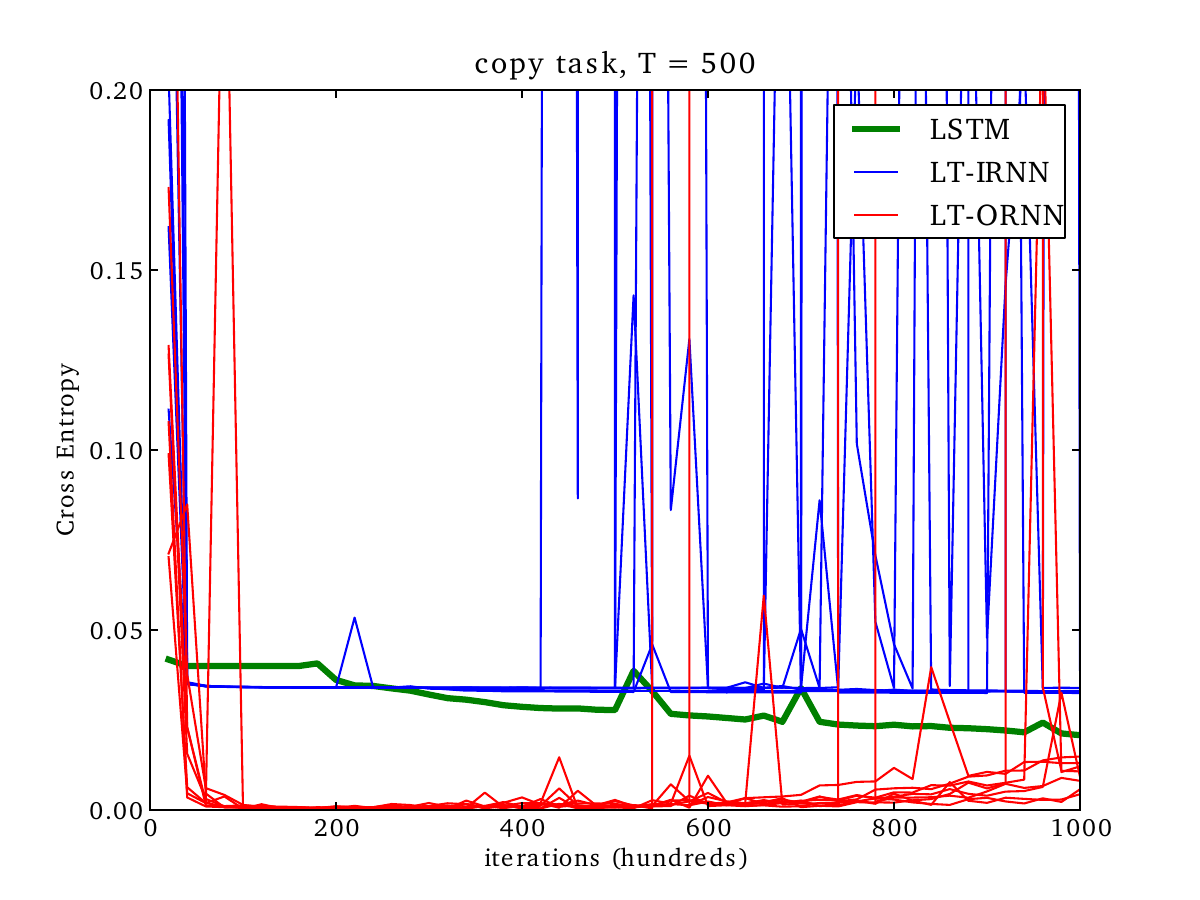}}
    \caption{Results for the copy task.}
    \label{fig:copy}
\end{figure*}

\begin{figure*}[t]
    \centering
    \subfigure{\includegraphics[width=0.65\columnwidth]{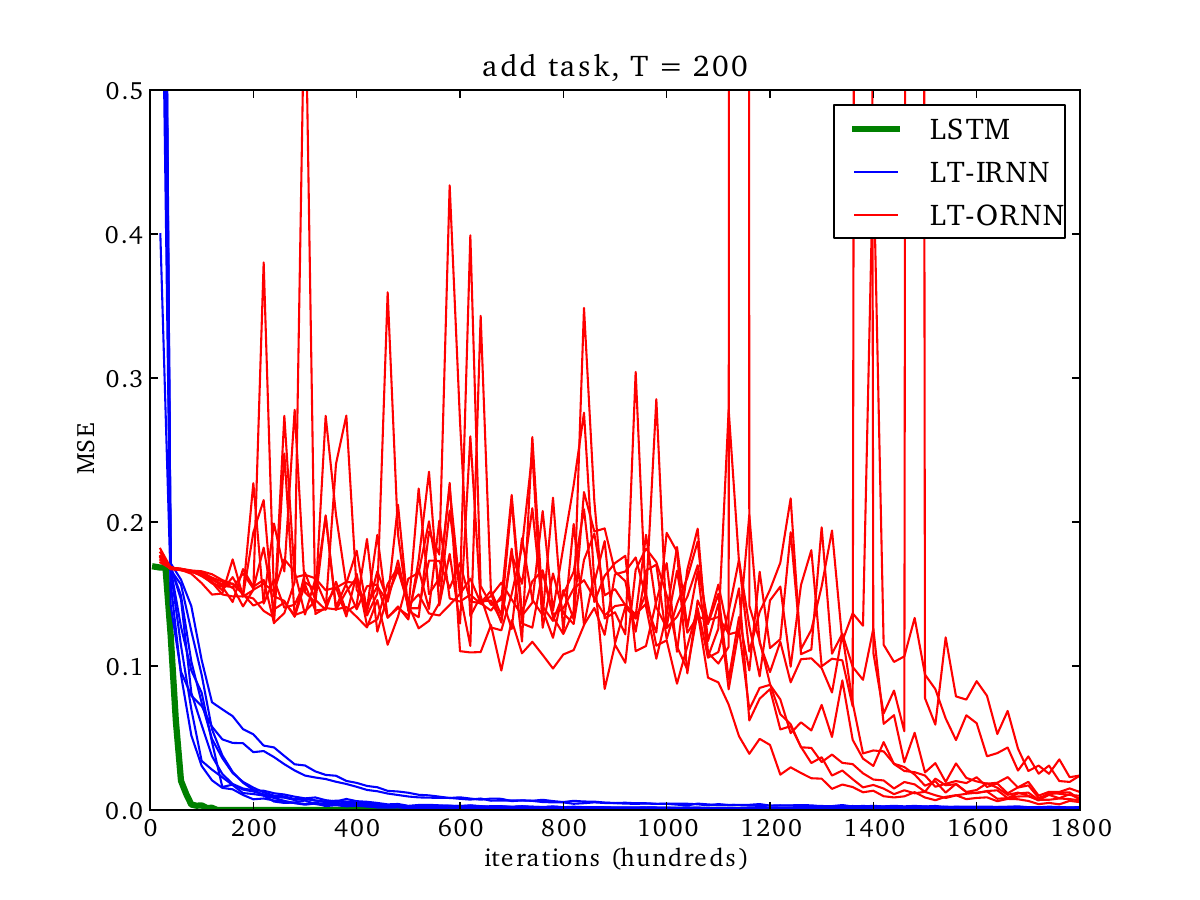}}\quad
    \subfigure{\includegraphics[width=0.65\columnwidth]{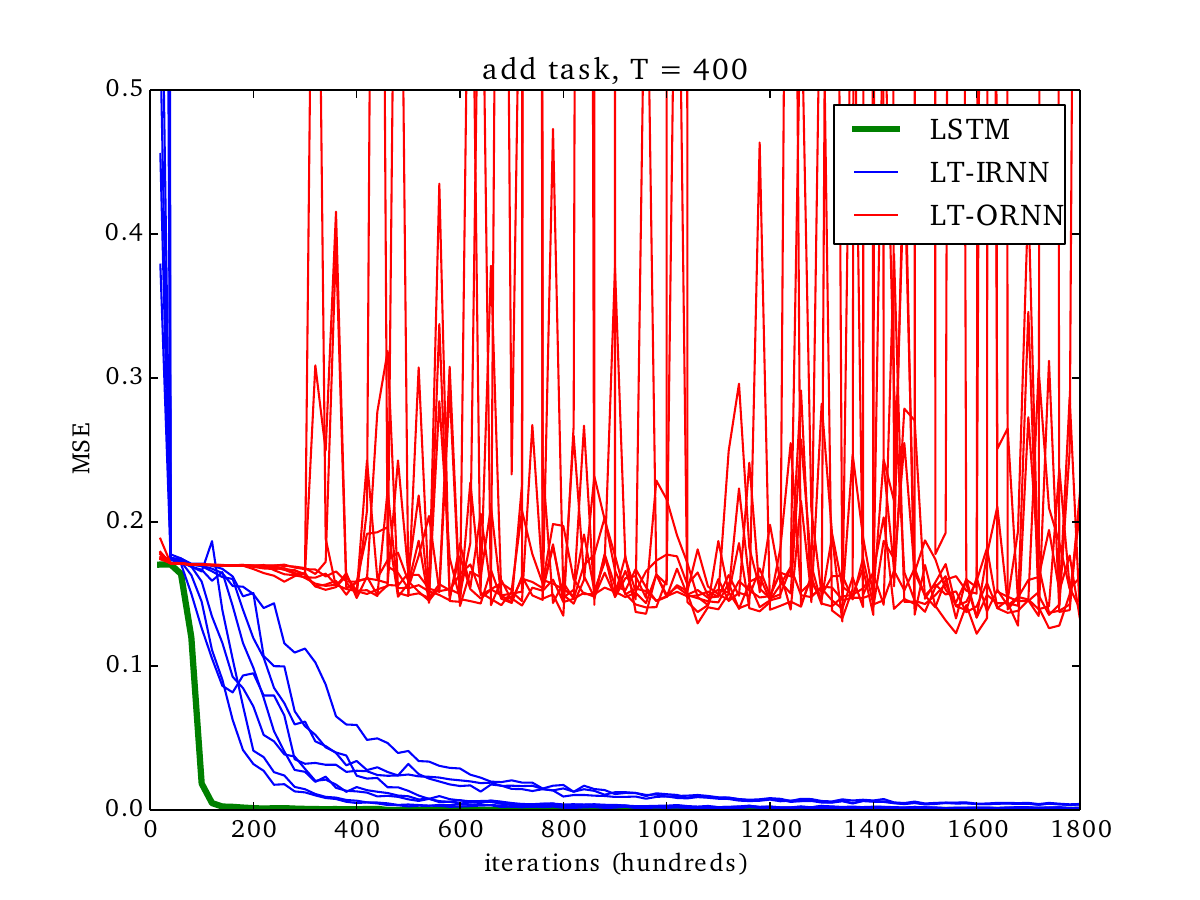}}
    \subfigure{\includegraphics[width=0.65\columnwidth]{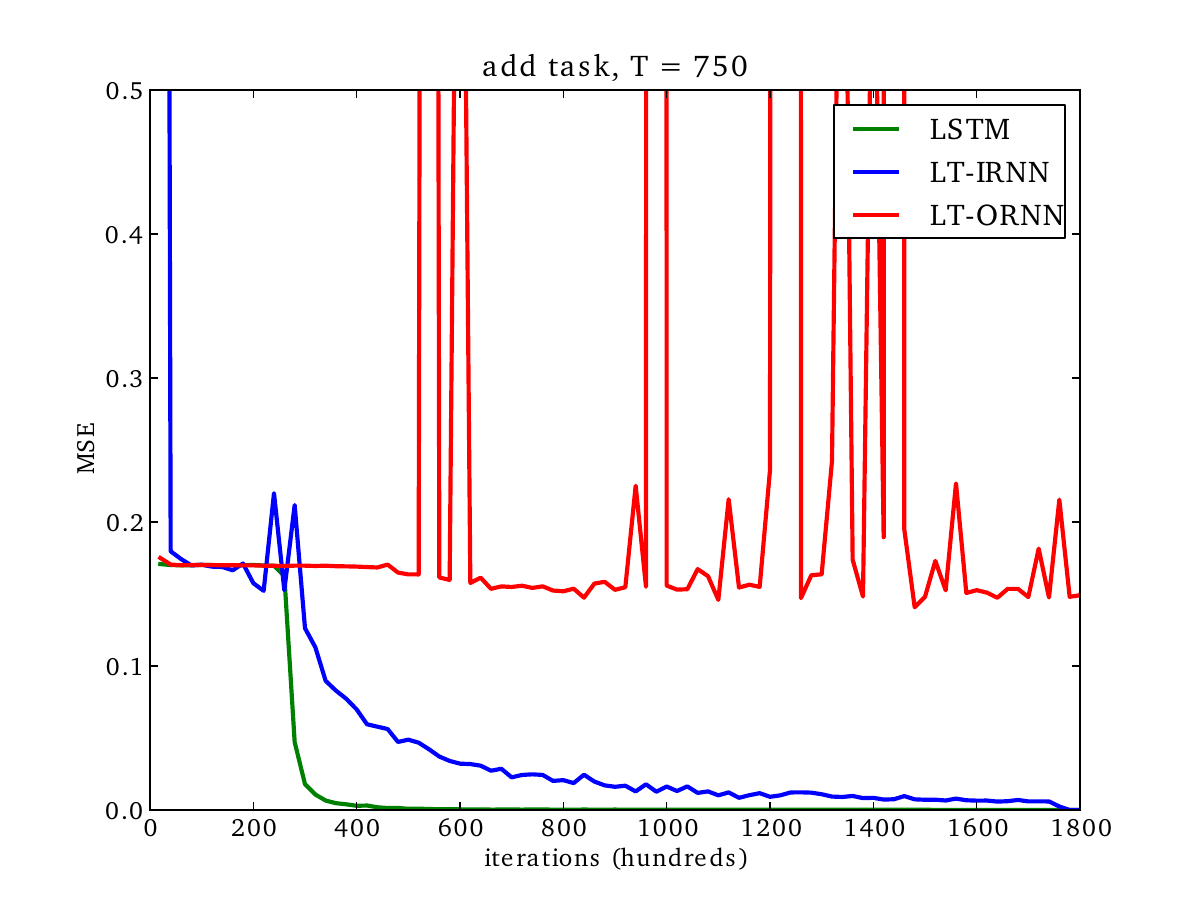}}
    \caption{Results for the addition task.}
    \label{fig:add}
\end{figure*}

\subsubsection{a solution mechanism}
This problem has a simple, explicit solution using a LT-RNN with a ReLU non-linearity and a one dimensional hidden state.  Namely:
set $U=[1 1]$, $b = -1$, $W=1$, and $V = 1$.  At each time step $j$, if $x_j[2] =  0$ then nothing is added to the hidden state, as
$x_j[1]\leq 1$.  On the other hand, if $x_j[2] =  1$, then exactly $x_j[2]$ is added to $h$.  

This mechanism has been known (at least implicitly, although we don't know if it has been written down explicitly before) at least since \cite{Hochreiter1997}, and it can be seen as a very simple LSTM model, with the following gates:
\begin{equation}
\begin{split}
i &= x_j[1] \\
f &= 1 \\
o &= 1 \\
g &= (x_j[2] - 1)_+  \\
\end{split}
\end{equation}
and no non-linearity in Equation (5).

\subsection{Comparison between the tasks}
Note that the $1\times 1$ matrix $V$ in the mechanism for the adding problem is the ``identity''.  We can build a more redundant solution by using a larger identity matrix.   We can describe the identity using the same block structure as the matrix $Q$ defined for the copy task; namely each $l_j=0$.    On the other hand, the $Q$ for the copy task acts as a ``clock'' that synchronizes after a fixed number of steps $T+S$.   It is important for the mechanism we described that the clock looks random at any time between $1$ and $S+T$.  For example, if we had instead used the same $l_j$ in each block $Q_j$, the mechanism would not succeed.  The transition matrices for the addition task and the copy task are thus opposites in the sense that for addition, all the $l_j$ are the same (i.e. a $\delta$ mass on the unit circle), and for copy, the $l_j$ are as uniformly distributed on the unit circle as possible.

In the experiments below, we will show that it is hard for an LT-RNN to learn the adding task when its transition matrix is initialized as a random orthogonal matrix but easy when initialized with the identity, and vice-versa for the copy task.  One way to get a ``unified'' solution is to use $l_2$ pooling, as in \ref{eq:pooled_RNN}.  Then when initialized with a matrix with $l_j$ distributed uniformly, the decoder can choose to use the pooled hiddens (which through away the phase, and so appear identity-like) for the adding task, or use the raw hiddens, which are clock-like.

\section{Experiments}

\subsection{Impact of Initialization}

Based on the above analysis, we hypothesize that an LT-RNN with random orthogonal initialization (denoted LT-ORNN) should perform well on the sequence memorization problem, and an LT-RNN with identity initialization (denoted LT-IRNN) should perform well on the addition task. 
To test this, we conducted the following experiment on both the copy and addition task for different timescales. 
For each task and timescale, we trained 8 LT-ORNNs and 8 LT-IRNNs with different random seeds.
The transformation matrices for all models were intialized using a Gaussian distribution with mean 0 and variance $1/\sqrt{n}$ (where $n$ is the number of incoming connections to each hidden unit). For LT-ORNNs, we then projected the transition matrix to its nearest orthogonal matrix by setting its singular values to 1.

In all experiments, we used RMSProp to train our networks with a fixed learning rate and a decay rate of $0.9$. 
In preliminary experiments we tried different learning rates in $\{1, 10^{-1}, 10^{-2}, 10^{-3}, 10^{-4},10^{-5}\}$
and chose the largest one for which the loss did not diverge, for the LT-RNN's we used  $10^{-4}$.

\begin{figure*}[t]
    \centering
    \subfigure{\includegraphics[width=1\columnwidth]{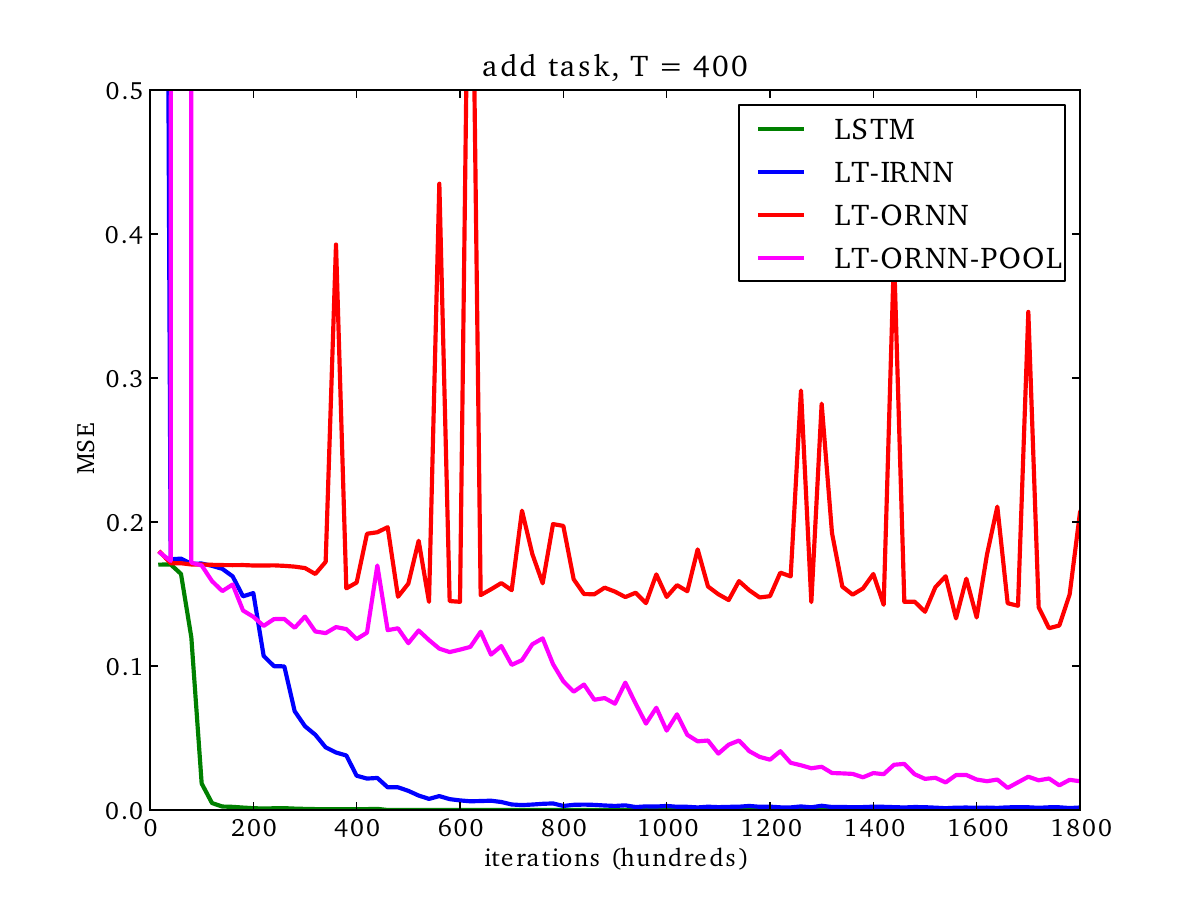}}\quad
    \subfigure{\includegraphics[width=1\columnwidth]{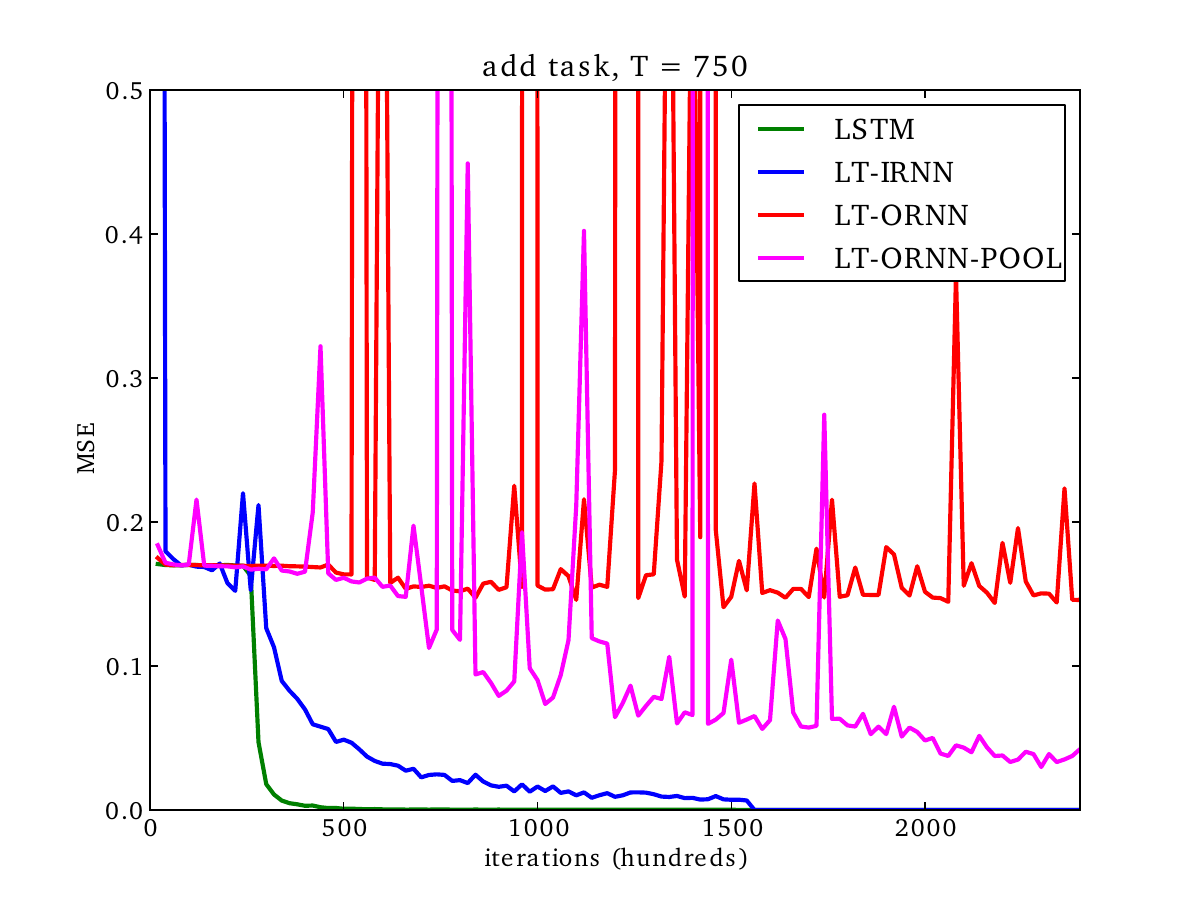}}
    \subfigure{\includegraphics[width=1\columnwidth]{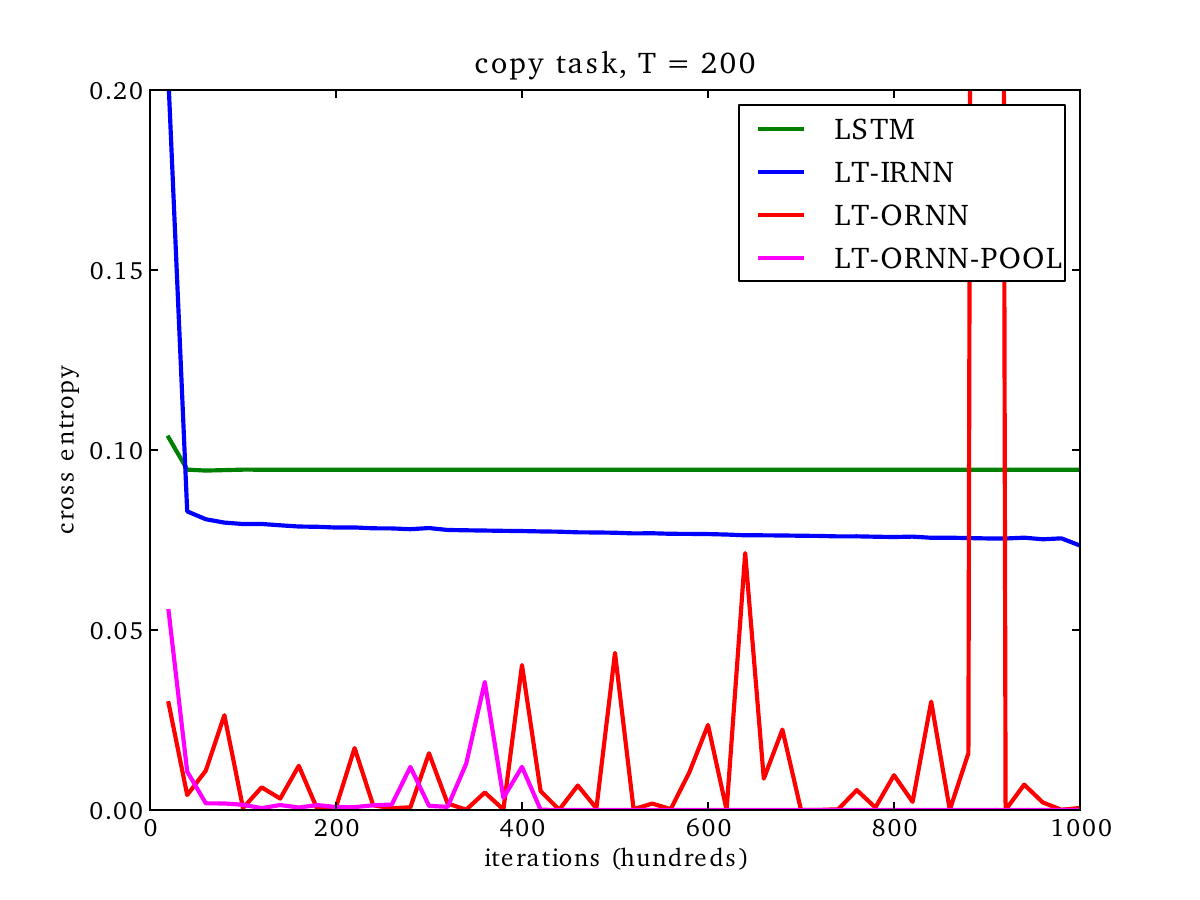}}\quad
    \subfigure{\includegraphics[width=1\columnwidth]{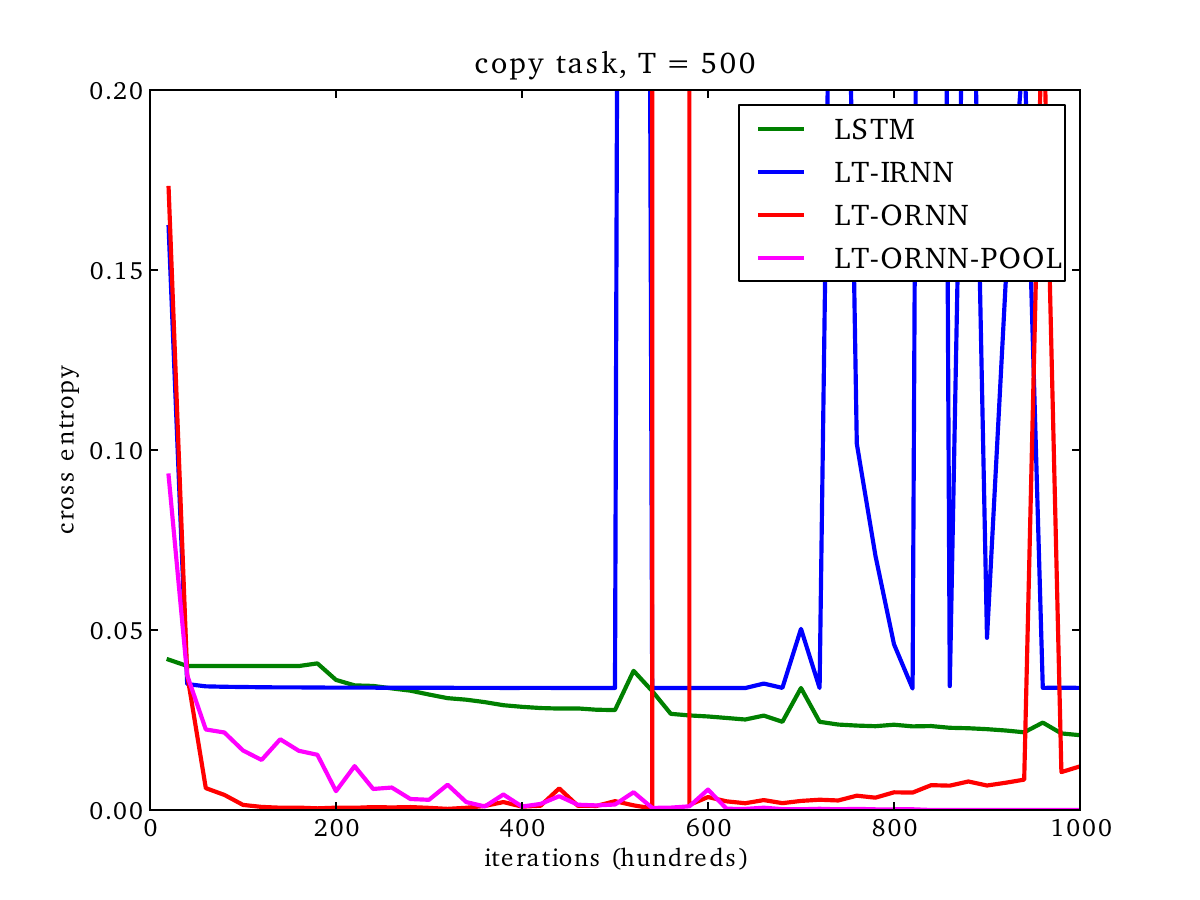}}
    \caption{Results for copy and addition task with pooling architectures.  Note that the LSTM will eventually solve the copy task, but the LT-IRNN will not.}
    \label{fig:pool}
\end{figure*}

We also included LSTMs in all our experiments as a baseline.   We used the same method as for LT-RNN to pick the learning rate, and ended up with $10^{-3}$.

For all experiments, we normalized the gradients with respect to hidden activations by $1/T$, where $T$ denotes the number of timesteps. 
In preliminary experiments, we also found that for LT-RNN models the activations frequently exploded whenever the largest singular value of the transition matrix became much greater than 1. Therefore, we adopted a simple activation clipping strategy where we rescaled activations to to have magnitude $l$ whenever their magnitude exceeded $l$. In our experiments we chose $l=1000$. 

Figure \ref{fig:copy} shows the results on the copy task for the LSTM, LT-ORNN and LT-IRNN. 
All networks are trained with 80 hidden units. 
We see that the LSTM has difficulty beating the baseline performance of only outputting the empty symbol; however it does eventually converge to the solution (this is not shown in the figure). 
However, the LT-ORNN solves the task almost immediately.
We note that this behavior is similar to that of the uRNN in \cite{Arjovsky2015}, which is paramaterized in a way that makes it easy to recover the explicit solution described above. 
The LT-IRNN is never able to find the solution. 

Figure \ref{fig:add} shows the results of the addition task for $T=200, 400$ and $750$ timesteps. 
All networks are trained with 128 hidden units. 
For $T=750$, we trained a single LT-ORNN and LT-IRNN due to time constraints. 
In contrast to the copy task, here the LT-IRNN is able to efficiently solve the problem whereas the LT-ORNN is only able to solve it after a very long time, or not at all. 
The LSTM is also able to easily solve the task, which is consistent with the original work of \cite{Hochreiter1997} where the authors report solving the task for up to 1000 timesteps. We note that this LSTM baseline differs from that of \cite{Arjovsky2015, Le2015} where it is reported to have more difficulty solving the addition task. We hypothesize that this difference is due to the use of different variants of the LSTM architecture such as peephole connections. 



\begin{figure*}[htp]
\centering
\begin{minipage}[b]{.45\textwidth}
\includegraphics[width=0.95\columnwidth]{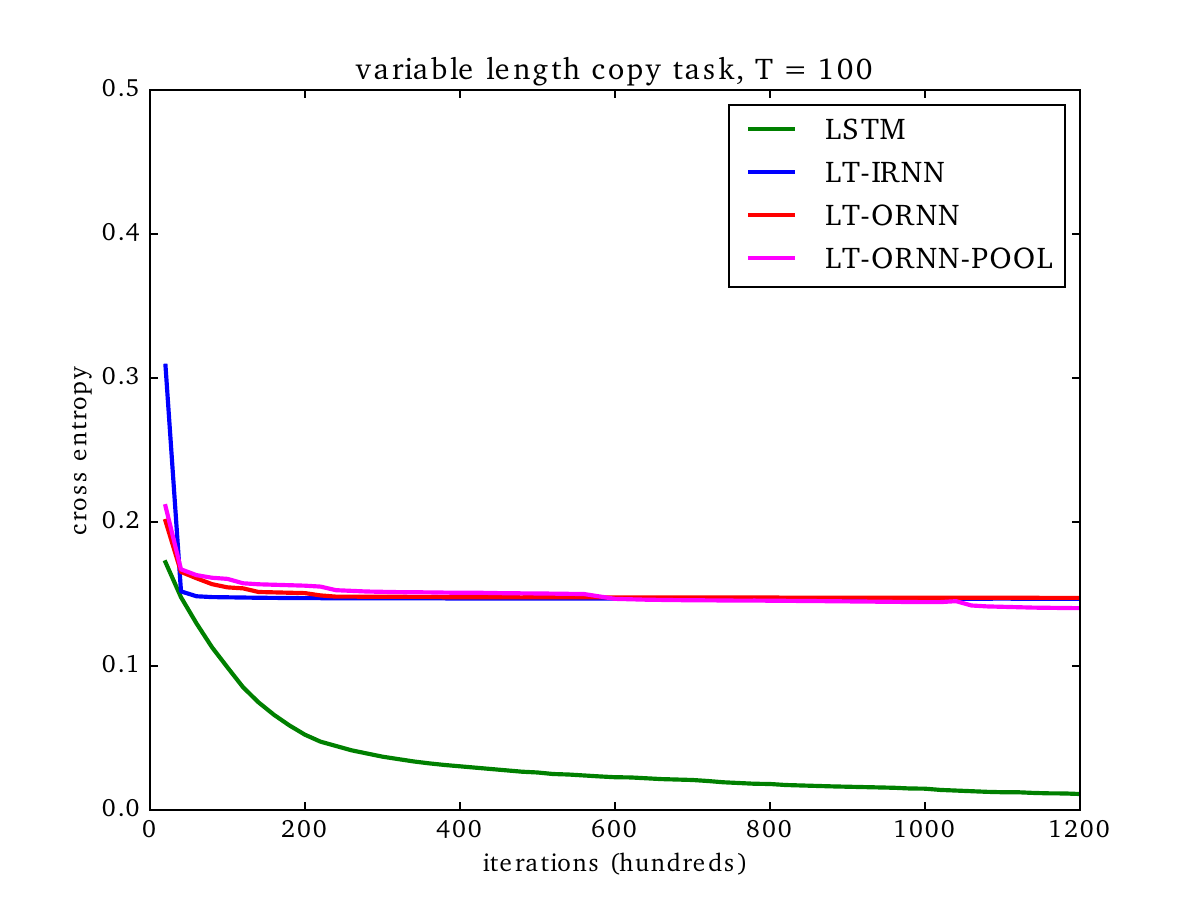}
\caption{Results for the variable length copy task.}\label{fig:random-regurg}
\end{minipage}\qquad
\begin{minipage}[b]{.5\textwidth}
\includegraphics[width=1.0\columnwidth]{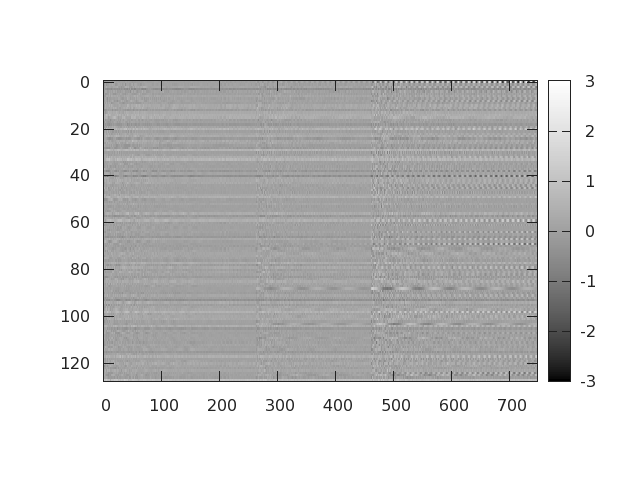}
\caption{Activation patterns of pooling network. The two marked numbers to be added occur at positions 264 and 463.}\label{fig:activations}
\end{minipage}
\end{figure*} 

\subsection{Pooling Experiments}

We next ran a series of experiments to examine the effect of feeding pooled outputs to the decoder, to see if we could obtain good performance on both the copy and addition tasks with a single architecture and initialization.
In these experiments, we added a soft penalty on the transition matrix $V$ to keep it orthogonal throughout training. 
Specifically, at every iteration we applied one step of stochastic gradient descent to minimize the loss $||V^TV - I ||$, evaluated at $m$ random points on the unit sphere. Note that this requires $\mathcal{O}(md^2)$ operations and a regular update requires $\mathcal{O}(Tmd^2)$ operations, so adding this soft constraint has negligible computational overhead. In our experiments we set $m=50$, which was the same at the minibatch size.

In all pooling experiments we used a pool size and stride of 2. 
The results are shown in Figure \ref{fig:pool}. 
The LT-ORNN with pooling is easily able to solve the copy task for both timescales, and approximately solves the addition task for both timescales as well, even though convergence is slower than the LT-IRNN. 
Its success on the copy task is not surprising, since by zeroing out the matrix $W_P$ in Equation \ref{eq:pooled_RNN} it can solve the problem with the same solution as the regular LT-ORNN.
The good performance on the adding task is somewhat more interesting.
To gain insight into how the network stores information in a stable manner while having an (approximately) orthogonal transition matrix, we plotted the activations of its hidden states over time as it processes an input sequence. 
This is displayed in Figure \ref{fig:activations}. 
We observe relatively constant activations until the first marked number is encountered, which triggers oscillatory patterns along certain dimensions. 
When the second marked number is seen, existing oscillations are amplified and new ones emerge. 
This suggests that the network stores information stably through the radius of its hidden state's rotations along different 2-dimensional subspaces.
The information is then recovered as the phase is discarded though the pooling operation.  Thus the model can have ``uniform'' clock-like oscillations that are perceived as $\delta$-like after the pooling.

\subsection{Variable Length Copy Task}

Having seen the stark impact of initialization on the performance of LT-IRNNs and LT-ORNNs for the copy and addition task, and its mitigation through the addition of a pooling layer, we then tested all the models on a problem for which we did not have a (roughly fixed size) solution mechanism, namely the variable length copy task.
Figure \ref{fig:random-regurg} shows the performance of an LT-IRNN, LT-ORNN, LT-ORNN with $l_2$ pooling, and LSTM (each with 80 hidden units) on the variable length copy task with $T=100$ timesteps.
Even though the number of timesteps is significantly less than in other tasks, none of the LT-RNNs are able to beat the chance baseline, whereas the LSTM is able to solve the task even though its convergence is slow. 
This experiment is a classic example of how a detail of construction of a synthetic benchmark 
can favor a model in a way that fails to generalize to other tasks.

\section{Conclusion}

In this work, we analyzed two standard synthetic long-term memory problems and provided explicit RNN solutions for them.  
We found that the (fixed length $T$) copy problem can be solved using an RNN with a transition matrix that is a $T+S$ root of the identity matrix $I$, and whose eigenvalues are well distributed on the unit circle, and we remarked that random orthogonal matrices almost satisfy this description.    
We also saw that the addition problem can be solved with $I$ as a transition matrix.  We showed that correspondingly, initializing with $I$ allows a linear-transition RNN to easily be optimized for solving the addition task, and initializing with a random orthogonal matrix allows easy optimization for the copy task; but that flipping these leads to poor results.
This suggests an optimization difficulty in transitioning between oscillatory and steady dynamics, which can be mitigated by adding an $l_2$ pooling layer that allows the model to easily choose between the two regimes.
Finally, our experiment with the variable length copy task illustrates that although synthetic benchmarks can be useful for evaluating specific capabilities of a given model, success does not necessarily generalize across different tasks, and novel model architectures should be evaluated on a broad set of benchmarks as well as natural data.

\nocite{langley00}

\bibliography{rnn_icml2016}
\bibliographystyle{icml2016}

\end{document}